\documentclass{article} 
\usepackage{helper_functions,times}
\usepackage{hyperref}
\usepackage{url}
\usepackage{graphicx}
\usepackage{amsmath}
\usepackage{caption}
\usepackage{subcaption}
\usepackage{float}
\usepackage[utf8]{inputenc}  
\usepackage[T1]{fontenc}  
\usepackage{natbib}

\title{Towards Deep Symbolic Reinforcement Learning}

\author{
	Marta Garnelo\\
	Department of Computing\\
	Imperial College London\\
	London, SW7 2AZ\\
	\texttt{garnelo@ic.ac.uk} \\
	\AND
	Kai Arulkumaran\\
	Department of Bioengineering\\
	Imperial College London \\
	London, SW7 2AZ \\
	\texttt{kailash.arulkumaran13@ic.ac.uk} \\
	\And
	Murray Shanahan \\
	Department of Computing\\
	Imperial College London \\
	London, SW7 2AZ \\
	\texttt{m.shanahan@ic.ac.uk} \\
}

\begin{document}

	\maketitle
	
	\begin{abstract}
		Deep reinforcement learning (DRL) brings the power of deep neural networks to bear on the generic task of trial-and-error learning, and its effectiveness has been convincingly demonstrated on tasks such as Atari video games and the game of Go. 
		However, contemporary DRL systems inherit a number of shortcomings from the current generation of deep learning techniques. 
		For example, they require very large datasets to work effectively, entailing that they are slow to learn even when such datasets are available.
		Moreover, they lack the ability to reason on an abstract level, which makes it difficult to implement high-level cognitive functions such as transfer learning, analogical reasoning, and hypothesis-based reasoning.
		Finally, their operation is largely opaque to humans, rendering them unsuitable for domains in which verifiability is important. 
		In this paper, we propose an end-to-end reinforcement learning architecture comprising a neural back end and a symbolic front end with the potential to overcome each of these shortcomings. 
		As proof-of-concept, we present a preliminary implementation of the architecture and apply it to several variants of a simple video game.
		We show that the resulting system -- though just a prototype -- learns effectively, and, by acquiring a set of symbolic rules that are easily comprehensible to humans, dramatically outperforms a conventional, fully neural DRL system on a stochastic variant of the game.
	\end{abstract}

	\section{Introduction}
	
	Deep reinforcement learning (DRL), wherein a deep neural network~\citep{lecun2015deep, schmidhuber2015deep} is used as a function approximator within a reinforcement learning system~\citep{sutton1998reinforcement}, has recently been shown to be effective in a number of domains, including Atari video games~\citep{mnih2015human}, robotics~\citep{levine2016end}, and the game of Go~\citep{silver2016mastering}. DRL can be thought of as a step towards instantiating the formal characterisation of universal artificial intelligence presented by Hutter {\citep{legg2007universal}, a theoretical framework for artificial general intelligence (AGI) founded on reinforcement learning. However, contemporary DRL systems suffer from a number of shortcomings. First, they inherit from deep learning the need for very large training sets, which entails that they learn very slowly. Second, they are brittle in the sense that a trained network that performs well on one task often performs very poorly on a new task, even if the new task is very similar to the one it was originally trained on. Third, they are strictly reactive, meaning that they do not use high-level processes such as planning, causal reasoning, or analogical reasoning to fully exploit the statistical regularities present in the training data. Fourth, they are opaque. It is typically difficult to extract a humanly-comprehensible chain of reasons for the action choice the system makes. Each of these shortcomings is an active area of research in the DRL community, where data efficient learning~\citep{assael2015data, gu2016continuous}, transfer learning~\citep{parisotto2015actor, arulkumaran2016classifying, barreto2016successor, rusu2016progressive}, planning \citep{guo2014deep, mnih2016strategic}, and transparency~\citep{zahavy2016Graying} are all hot topics.
		
		Here we take a different approach. We propose a novel reinforcement learning architecture that addresses all of these issues at once in a principled way by combining neural network learning with aspects of classical symbolic AI, gaining the advantages of both methodologies without their respective disadvantages. Central to classical AI is the use of language-like propositional representations to encode knowledge. Thanks to their compositional structure, such representations are amenable to endless extension and recombination, an essential feature for the acquisition and deployment of high-level abstract concepts, which are key to general intelligence~\citep{mccarthy1987generality}. Moreover, knowledge expressed in propositional form can be exploited by multiple high-level reasoning processes and has general-purpose application across multiple tasks and domains. Features such as these, derived from the benefits of human language, motivated several decades of research in symbolic AI.
		
		\begin{figure}
			\centering
			\includegraphics[width=\textwidth]{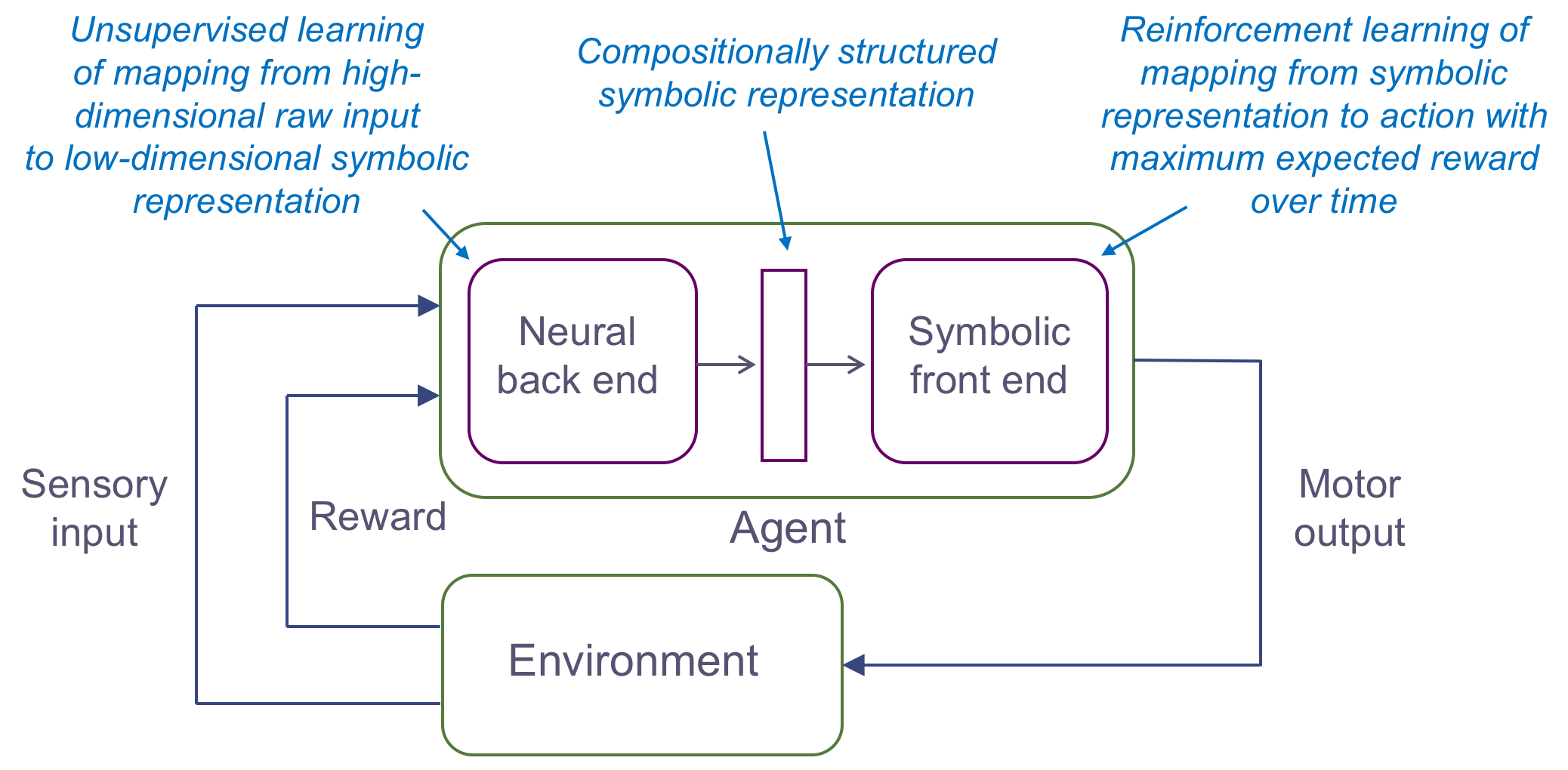}
			\label{fig:fig1}
			\caption{The proposed reinforcement learning architecture. The neural back end learns to map raw sensor data into a symbolic representation, which is used by the symbolic front end to learn an effective policy.}
		\end{figure}
		
		But as an approach to general intelligence, classical symbolic AI has been disappointing. A major obstacle here is the symbol grounding problem~\citep{harnad1990symbol, shanahan2005perception}. The symbolic elements of a representation in classical AI -- the constants, functions, and predicates -- are typically hand-crafted, rather than grounded in data from the real world. Philosophically speaking, this means their semantics are parasitic on meanings in the heads of their designers rather than deriving from a direct connection with the world. Pragmatically, hand-crafted representations cannot capture the rich statistics of real-world perceptual data, cannot support ongoing adaptation to an unknown environment, and are an obvious barrier to full autonomy. By contrast, none of these problems afflict machine learning. Deep neural networks in particular have proven to be remarkably effective for supervised learning from large datasets using backpropagation~\citep{lecun2015deep, schmidhuber2015deep}. Deep learning is therefore already a viable solution to the symbol grounding problem in the supervised case, and for the unsupervised case, which is essential for a full solution, rapid progress is being made~\citep{chen2016infogan, goodfellow2014generative, greff2016tagger, higgins2016early, kingma2013auto}. The hybrid neural-symbolic reinforcement learning architecture we propose relies on a deep learning solution to the symbol grounding problem.
		
		At the top level, the deep symbolic reinforcement learning architecture we propose is very simple (Fig. 1). It comprises a deep neural network back end, whose job is to transform raw perceptual data into a symbolic representation, which is fed to a symbolic front end whose task is action selection. The function computed by each half of the architecture is shaped by machine learning, so the system as a whole learns end-to-end with minimal assumptions made on the nature of the environment. The neural back end must learn a compositionally-structured compressed representation of the raw perceptual data, while the symbolic front end must learn a mapping from the resulting symbolic representation to actions that maximise expected reward over time.
		
		In this paper we present one instantiation of this architecture as a proof-of-concept, and illustrate its effectiveness on several variants of a simple video game. This demonstrator system has many limitations and makes numerous simplifying assumptions that are not inherent in the larger proposal, but it illustrates the four fundamental principles of our architectural manifesto. (For a related set of desiderata see~\citep{lake2016building}.)
		
		1) \textbf{Conceptual abstraction}. Determining that a new situation is similar or analogous to one (or several) encountered previously is an operation fundamental to general intelligence, and to reinforcement learning in particular. In a conventional DRL system, such as DQN~\citep{mnih2015human}, this is achieved through the generalising capabilities of the neural network that approximates the Q function (or the value function or policy function, depending on the style of reinforcement learning in question). However, this low-level approach to establishing similarity relationships requires the gradual build-up of a statistical picture of the state space. The upshot is that while a novice human player will rapidly spot the high-level similarity between, say, the paddle and ball in Pong and the paddle and ball in Breakout, a conventional DRL system is blind to this. By contrast, the present architecture maps high-dimensional raw input into a lower-dimensional conceptual state space within which it is possible to establish similarity between states using symbolic methods that operate at a higher level of abstraction. This facilitates both data efficient learning and transfer learning as well as providing a foundation for other high-level cognitive processes such as planning, innovative problem solving, and communication with other agents (including humans).
		
		2) \textbf{Compositional structure}. To enable this sort of conceptual abstraction, a representational medium is required that has a compositional structure. That is to say it should comprise a set of elements that can be combined and recombined in an open-ended way. Classically, the theoretical foundation for such a representational medium is first-order logic, and the underlying language comprises predicates, quantifiers, constant symbols, function symbols, and boolean operators~\citep{mccarthy1987generality}. (It should be noted that a fixed-size vector representation is inadequate for such a representational medium, because it can encode formulae of arbitrary length.) But the binary nature of classical logic makes it less well suited to dealing with the uncertainty inherent in real data than a Bayesian approach. To handle uncertainty, we propose probabilistic first-order logic for the semantic underpinnings of the low-dimensional conceptual state space representation into which the neural front end must map the system’s high-dimensional raw input~\citep{halpern1990analysis}.
		
		3) \textbf{Common sense priors}. Although our target is general intelligence, meaning the ability to achieve goals and perform tasks in a wide variety of domains, it is unrealistic to expect an end-to-end reinforcement learning system to succeed with no prior assumptions about the domain. For example, in most DRL systems that take visual input, spatial priors, such as the likelihood that similar 2D patterns will appear in different locations in the visual field, are implicit in the convolutional structure of the network~\citep{bengio2013representation}. But the everyday physical world is structured according to many other common sense priors~\citep{hayes1985second, bengio2013representation, davis2014representations, lake2016building, mueller2014commonsense}. Consisting mostly of empty space, it contains a variety of objects that tend to persist over time and have various attributes such as shape, colour, and texture~\citep{shanahan1995default}. Objects frequently move, typically in continuous trajectories. Objects participate in a number of stereotypical events, such as starting to move or coming to a halt, appearing or disappearing, and coming into contact with other objects. These minimal assumptions and expectations can be built into the system by grafting a suitable ontology onto the underlying representational language, greatly reducing the learning workload and facilitating various forms of common sense reasoning.
		
		4) \textbf{Causal reasoning}. The current generation of DRL architectures eschews model-based reinforcement learning, ensuring that the resulting systems are purely reactive. By contrast, the architecture we propose attempts to discover the causal structure of the domain, and to encode this as a set of symbolic causal rules expressed in terms of the common sense ontology described above. These causal rules enable conceptual abstraction. As already mentioned, the key to general intelligence is the ability to see that an ongoing situation is similar or analogous to a previously encountered situation or set of situations. A deep neural network that approximates the Q function in reinforcement learning can be thought of as carrying out analogical inference of this kind, but only at the most superficial, statistical level. To carry out analogical inference at a more abstract level, and thereby facilitate the transfer of expertise from one domain to another, the narrative structure of the ongoing situation needs to be mapped to the causal structure of a set of previously encountered situations. As well as maximising the benefit of past experience, this enables high-level causal reasoning processes to be deployed in action selection, such as planning, lookahead, and off-line exploration (imagination).
		
		Our implemented proof-of-concept system embodies each of these principles, albeit in a restricted form. The back end of the system learns to construct symbolic representations of sequences of game states, in which the flow of raw pixel data is encoded in a more conceptually abstract form, defined in terms of objects, their types, locations, and interactions. These representational elements can be joined in arbitrary combinations, yielding compositional structure. The ontology of this representational medium reflects common sense priors such as the tendency of objects to persist over time and the default assumption that objects that look alike behave in similar ways. Finally, the symbolic front end of the system learns effective policies because, in a rudimentary sense that is nevertheless inapplicable to conventional DRL systems, it ``understands'' the causal relations between its own actions, interactions among objects, and the acquisition of reward. However, in the present prototype the neural back end is in fact relatively shallow, while the symbolic front end carries out very little high-level reasoning. Consequently, it barely scratches the surface of what we believe is possible with the architecture we are proposing, and it should be regarded as just a first step towards a general purpose AI system that fully exploits the combined power of deep learning and symbolic reasoning.

		\section{Experimental setup}
		
		As a benchmark for our prototype system, we implemented several variants of a simple game where the agent (shaped as a `+') has to learn either to avoid or to collect objects depending on their shape. Once the agent reaches an object using one of four possible move actions (up, down, left, or right), this object disappears and the agent obtains either a positive or a negative reward. Encountering a circle (`o') results in a negative reward while collecting a cross (`x') yields a positive reward.
		
		We applied the system to four variants of this game (figure~\ref{fig:screens}) wherein the type of objects involved and their initial positions are as follows:
		
		\textbf{Variant 1.} In this environment there are only objects that return negative rewards (`o') and they are positioned in a grid across the screen. This layout is the same for every new game. Encountering an object returns a score of -1 and at the beginning of the game the player is located in the middle of the board.
		
		\textbf{Variant 2.} The layout is the same as in version 1 but there are two types of objects. As before, circles give -1 points and we introduce crosses that return 1 points.
		
		\textbf{Variant 3.} As in version 1 this game only contains objects that return a negative reward. In order to increase the difficulty of the learning process however, the position of these objects is determined at random and changes at every new game.
		
		\textbf{Variant 4.} This version combines the randomness from environment 3 and the different object types from version 2.

		\begin{figure}
			\centering
			\label{fig:games}
			\begin{subfigure}[b]{0.245\textwidth}
				\includegraphics[width=\textwidth]{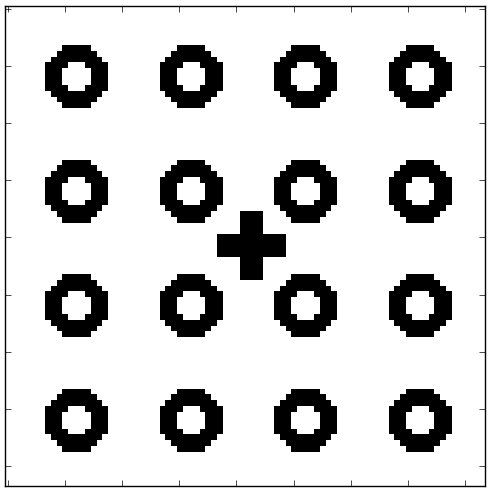}
				\label{fig:game1}
			\end{subfigure}
			\begin{subfigure}[b]{0.245\textwidth}
				\includegraphics[width=\textwidth]{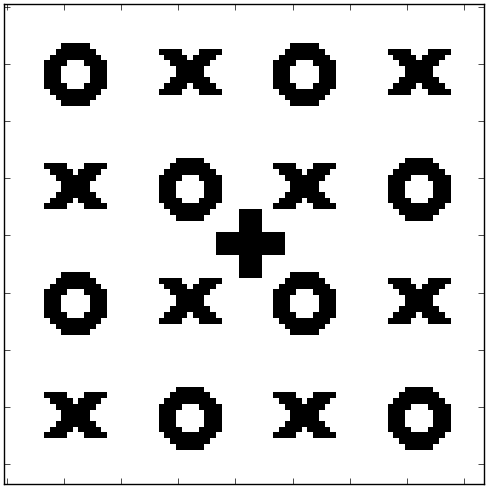}
				\label{fig:game2}
			\end{subfigure}
			\begin{subfigure}[b]{0.245\textwidth}
				\includegraphics[width=\textwidth]{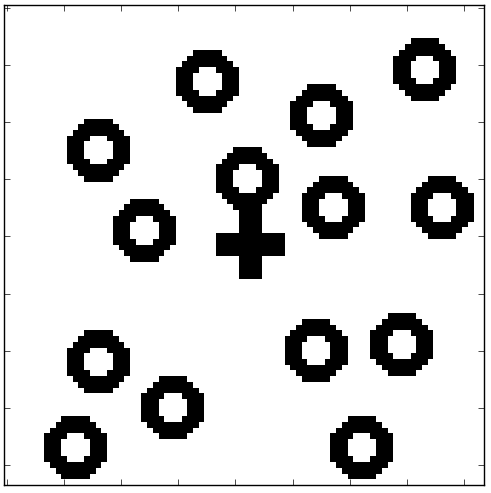}
				\label{fig:game3}
			\end{subfigure}
			\begin{subfigure}[b]{0.245\textwidth}
				\includegraphics[width=\textwidth]{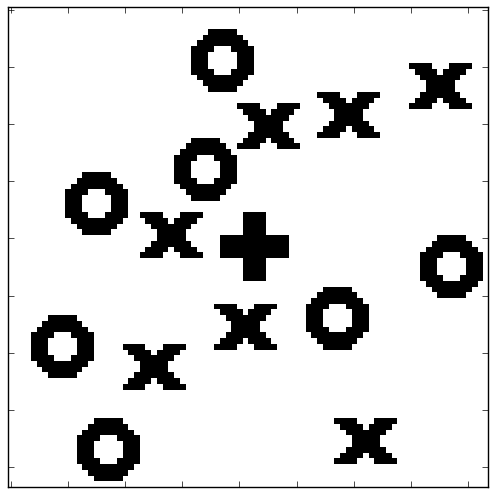}
				\label{fig:game4}
			\end{subfigure}
			\caption{The four different game environments. The agent is represented by the `+' symbol. The static objects return positive or negative reward depending on their shape (`x' and `o' respectively).}
			\label{fig:screens}
		\end{figure}

		\section{Methods}
		
		Our algorithm can be thought of as a pipeline comprising three stages: low-level symbol generation, representation building, and reinforcement learning.

		\subsection{Low-level symbol generation}
		
		The goal of this first stage is to generate, in an unsupervised manner, a set of symbols that can be used to represent the objects in a scene. We use a convolutional neural network for this, since such networks are well-suited to feature extraction, especially from images. Specifically, we train a convolutional autoencoder on 5000 randomly generated images of varying numbers of game objects scattered across the screen.
		
		Given the simplicity of the images, which consist of objects with three different geometric shapes (cross, plus sign, and circle) on a uniform background, we don't need to train a deep network to detect the different features for our current game benchmark. Our network consists of a 5x5 convolutional layer followed by a 2x2 pooling layer plus the corresponding decoding layers. The activations across features in the middle layer of the CNN are used directly for the detection of the objects in the scene.

		\paragraph{Object detection and characterisation.} The next step of the symbol generation stage uses the salient regions of the convoluted image (figure~\ref{fig:fig_2}).
		As shown by Li et al~\citep{li2016relief}, salient areas in an image will result in higher activations throughout the layers of a convolutional network. 
		Given that our games are geometrically very simple, this property is enough to enable the extraction of the individual objects from any given frame. 
		To do this, we first select, for each pixel, the feature with the highest activation. We then threshold these activation values, forming a list of those that are sufficiently salient. Ideally, each member of this list is a representative pixel for a single object.
		The objects identified this way are then assigned a symbolic type according to the geometric properties computed by the autoencoder. 
		This is done by comparing the activation spectra of the salient pixels across features. 
		In the current implementation, this comparison is carried out using the sum of the squared distances, which involves setting an ad hoc threshold for the maximal distance between two objects of the same type. In future work with richer environments, we anticipate using more sophisticated clustering algorithms as the state of the art in unsupervised learning advances.
		
		The information extracted at this stage consists of a symbolic representation of the positions of salient objects in the frame along with their types. (See the left-hand table of figure~\ref{fig:fig_e}.)
		
		\begin{figure}
			\centering
			\includegraphics[width=\textwidth]{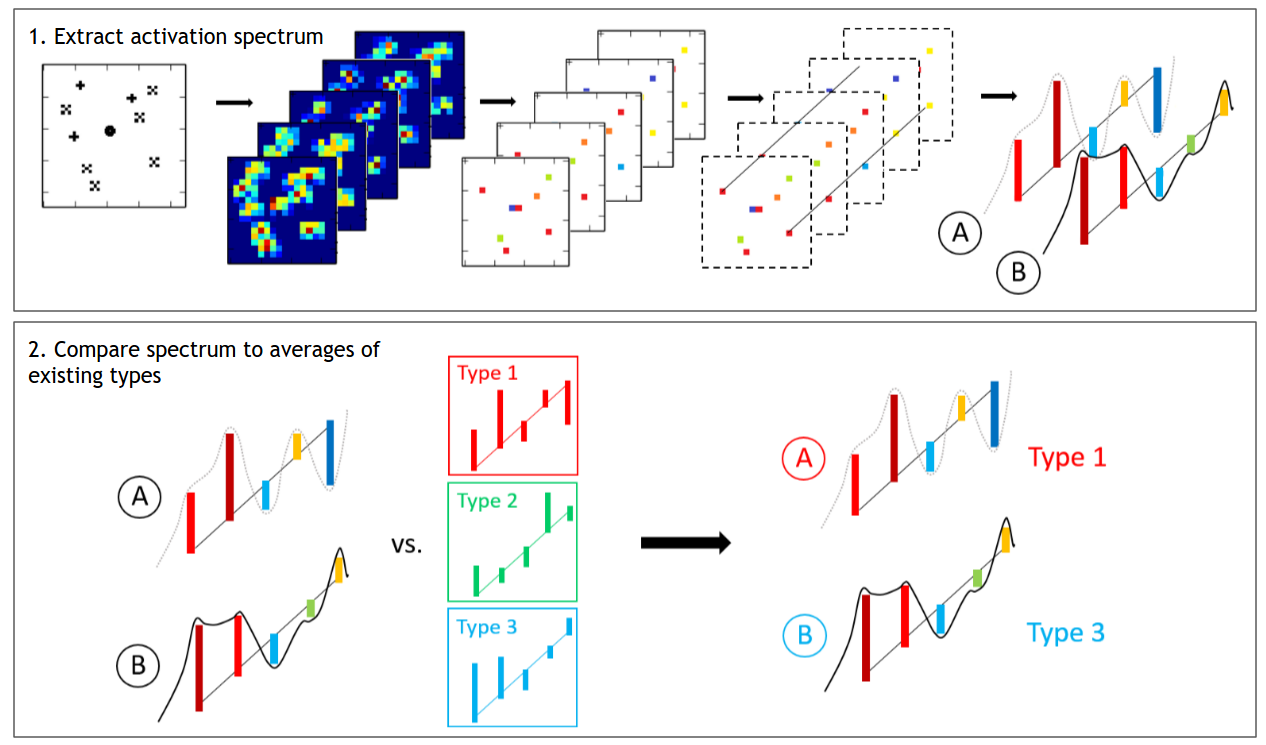}
			\caption{Unsupervised extraction of low-level symbols from the information provided by the convolutional autoencoder. The spectrum of activations across features is obtained by selecting the pixels with the highest activations across features. In a second step each symbol is assigned one of the existing types by comparing their spectra. If no match is found a new type is created.}
			\label{fig:fig_2}
		\end{figure}

		\subsection{Representation building}
		
		Once we have extracted the low-level symbols from a single snapshot of the game we need to be able to track them across frames in order to observe and learn from their dynamics. To do this, we need to take account of the first common sense prior: object persistence across time.
		The concept of persistence we deploy is based on three measures combined into a single value. This value, which captures the likelihood that object~$i_1^t$ identified in one frame is the same as object~$i_2^{t+1}$ identified in the next frame, is a function of:
		
		\paragraph{Spatial proximity.} We build in the notion of continuity by defining the likelihood to be inversely proportional to the distance between two objects in consecutive frames. We have
		\begin{equation*}
			L_{dist} = \frac{1}{1+d}
		\end{equation*}
		where d is the Euclidean distance between two objects $i_1^t$ and $i_2^{t+1}$ in consecutive frames $t$ and $t+1$ respectively. Distance, however, is not the sole determiner of an object's identity, given that objects can replace each other.
		
		\paragraph{Type transitions.} Given the types of two objects $\tau(i_1^t) = \tau_{i1}$ and $\tau(i_2^{t+1}) = \tau_{i2}$ in consecutive frames, we can determine the probability that they are the same object that has changed from one type to the other by learning a transition matrix $T$ from previously observed frames. Given that this is a transition matrix it is already normalised to be a probability.
		
		\begin{equation*}
			L_{trans} = T_{\tau_{i1},\tau_{i2}}
		\end{equation*}
		
		In order to be able to describe all transitions, including the ones that correspond to objects appearing and disappearing, we introduce the object type 0. This type corresponds to 'non-existent'. An object of type 1 that appears in frame $t$ has thus carried out the transition $0 \to 1$. If that object disappears later on it would transition $1 \to 0$. This addition can only be carried out after the tracking step and once every object has been assigned its corresponding type in the previous time steps.
		
		\paragraph{Neighbourhood.} The neighbourhood of an object will typically be similar from one frame to the next. This allows us to discriminate between objects that are spatially close but approaching from different directions. For now we just consider the difference in the number of neighbours, $\Delta N$ between two objects, where we define a neighbour to be any object, $i_n$, within a distance $d_{max}$ of another object $i_1$. Future improvements could include considering the types of neighbours rather than just the number.
		
		\begin{equation*}
			L_{neigh} = \frac{1}{1+\Delta N}
		\end{equation*}
		
		We track an object by combining the three measures just described. We have
		
		\begin{equation*}
			L = w_1 L_{dist} + w_2 L_{trans} + w_3 L_{neigh}
		\end{equation*}
		
		where $w_1, ..., w_3$ are ad hoc weights. In future implementations these weights could be learned and set dynamically as the importance of each term varies during the learning process.
		
		\begin{figure}
			\centering
			\includegraphics[width=0.80\textwidth]{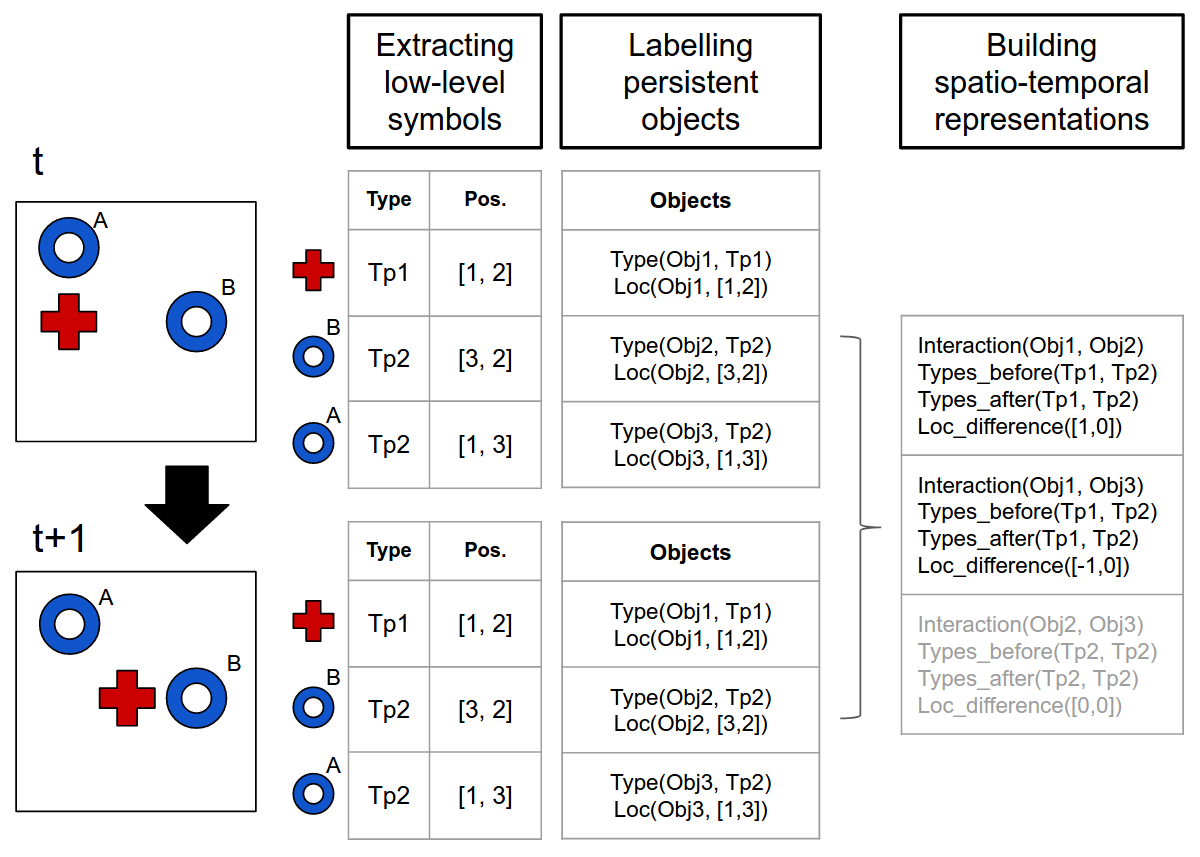}
			\caption{Overview of the symbolic representations extracted after each of the three main steps. The images on the left show two consecutive frames of a toy example. The first column corresponds to the information obtained from these frames after low-level symbol extraction from the convolutional autoencoder. After tracking, a unique label is assigned to each persistent object (Obj1-3) and its associated type and position is recorded. The last column on the right lists the interactions between objects which are used for the reinforcement learning stage. The information is now expressed in terms of relative distance rather than absolute position and corresponds to the difference between two frames.}
			\label{fig:fig_e}
		\end{figure}
		
		As a result of the tracking process, we acquire an additional attribute for each object in the scene: a unique identifier that labels it across time. This identifier is added to the symbolic representation of the ongoing game state. (See the middle table of figure~\ref{fig:fig_e}.)
		
		\paragraph{Symbolic interactions and dynamics.}
		So far, information about the game is expressed in terms of static frame-by-frame representations (albeit taking advantage of information in consecutive frames). But the final, reinforcement learning stage of the algorithm will require information about the dynamics of objects and their spatial interactions. This is obtained from the static representations constructed so far using two principles. First, we consider the difference between frames rather than working with single frames, thus moving to a temporally extended representation. Second, we represent the positions of objects relative to other objects rather than using absolute coordinates. Moreover, we only record relative positions of objects that lie within a certain maximum distance of each other. This approach is justified by the common sense prior that local relations between multiple objects are more relevant than the global properties of single objects.
		
		Of course, this assumption is not always valid, even with this game. As a result, the present system can arrive at a locally optimal policy that is inferior to the global optimum.
		So future implementations will have to handle this question of circumscribing relevance in a more nuanced way~\citep{shanahan2016frame}.
		But in general, the representational sparseness that results from adopting a locality assumption compensates for the loss of global optimality by allowing much faster training.
		
		We now have a concise spatio-temporal representation of the game situation, one that captures not only what objects are in the scene along with their locations and types, but also what they are doing. In particular, it represents frame-to-frame interactions between objects, and the changes in type and relative position that result. (See the right-hand table in figure~\ref{fig:fig_e}.) This is the input to reinforcement learning, the third and final stage of the pipeline.

		\subsection{Reinforcement learning}
		
		The spatio-temporal representation constructed in stages one and two of the system pipeline can now be used to learn an effective policy for game play. At this point, the advantage of using the locality heuristic becomes clear. A representation that included all possible relations between objects would result in a very large state space and very long training times, with most states being visited infrequently. However, most interactions between objects are independent of each other, both in the real world and in this particular game. So instead of representing all relations in one global state, our representation comprises a set of localised representations. This results in a significantly reduced state space and enables fast generalisation across object types.
		
		In order to implement this independence we train a separate Q function for each interaction between two object types. The main idea is to learn several Q functions for the different interactions and query those that are relevant for the current situation. Given the simplicity of the game and the reduced state space that results from the sparse symbolic representation we can approximate the optimal policy using tabular Q-learning. The update rule for the interaction between objects of types $i$ and $j$ is therefore 
		
		\begin{equation*}
			Q^{ij}(s^{ij}_t, a_t) 	\leftarrow Q^{ij}(s^{ij}_t, a_t) + \alpha \left( r_{t+1} + \gamma (\max_a Q^{ij}(s^{ij}_{t+1}, a) - Q^{ij}(s^{ij}_t, a_t)) \right)
		\end{equation*}
		
		where $\alpha$ is the learning rate, $\gamma$ is the temporal discount factor, and each state $s^{ij}_t$ represents an interaction between object types $i$ and $j$ at time step $t$. In this case the interactions are changes in relative distance between the objects in question. The state space $S^{ij}$ thus describes the different possible relations between two objects of types $i$ and $j$. Given that we have limited the interactions to a certain radius of proximity this state space is bounded but learning is not guaranteed to converge on a global optimum. Finally, in order to choose the next action we add up all Q values obtained from the currently relevant Q functions at the time step and pick the one that will return the highest reward overall.
		
		\begin{equation*}
			a_{t+1} = \arg\max_a ( \sum_Q(Q(s_{t+1}, a))
		\end{equation*}
		
		During training we use an $\epsilon$-greedy exploration strategy with a 10\% chance of choosing a different action at random.
		
		Note that, in the present benchmark games, the only moving object is the one the agent acts on directly (the plus sign), which ensures that every currently relevant Q function pertains to a possible agent action. In a more general implementation it will be necessary to automatically identify which objects in the world the agent controls directly (its own body, for example) and restrict the Q function accordingly.

		\section{Results}
		
		Agents were trained in epochs of 100 time steps for a maximum of 1000 epochs. We trained 20 agents separately and tested them for 200 time steps on 10 games at every tenth epoch. The resulting average score is plotted in figure~\ref{fig:LR} for all four games. In all four cases the score increases within the first few hundred epochs and remains approximately constant for the remaining time.
		
		While plotting the average score in this way is a common way of visualising successful learning for this type of experiment, this measure can produce an incomplete characterisation of the learning process in environments with both positive and negative objects. 
		There are two reasons for this. First, given that the scores returned by the objects can cancel each other out, there is more than one way to achieve any given final score. 
		For example, collecting ten positive and nine negative objects will result in the same final score as only collecting one positive object.
		Yet in the first case the number of positive objects the agent collects is about 53\% while in the second scenario it is 100\%. 
		Therefore, although they both have the same score, this percentage measure reveals that the agent in the latter case has learned to react to the different objects correctly while the agent in the former case collects items without regard for their type.
		Second, our agent only has a limited radius of view and can therefore get stuck in a location surrounded by negative objects.
		In this case the agent is forced to spend most of the test time avoiding these negative objects and won't obtain a high score.
		For these reasons we introduce a second measure: the percentage of positive objects collected of the total amount of objects encountered. 
		Rather than measuring how well our agent performs on a global level, this measure shows whether or not the agent has learned to interact correctly with the individual objects at a local level.
		
		The results for the two game variants that feature objects of two different types are shown in figure~\ref{fig:perc}. As expected, about 50\% of the objects that the agent initially collects are positive. As training continues the percentage increases to approximately 70\% in both cases.
		
		Finally, we tested the transfer learning capabilities of our algorithm by training an agent only on games of the grid variant then testing it on games of the random variant. While this setup is similar to the experiments on the random game in the sense that the grid setup can be seen as one among the possible random initialisations, the difference lies in the fact that the agent is exposed to just this one type of environment during training, whereas in the random experiments the agent experiences numerous random variations. As shown in figure~\ref{fig:LR_2}, the learning curve is comparable to the random case, albeit showing a slightly inferior performance.

		\begin{figure}
			\centering
			\begin{subfigure}[b]{0.45\textwidth}
				\includegraphics[width=\textwidth]{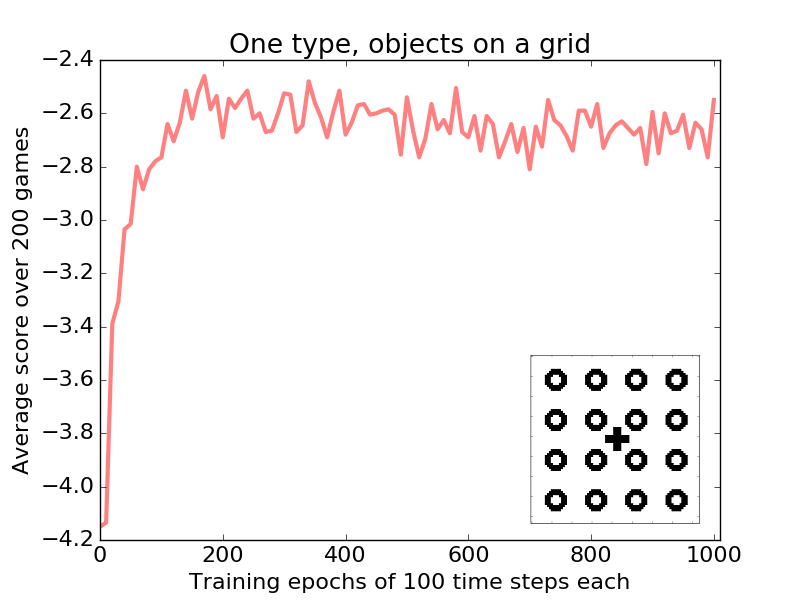}
			\end{subfigure}
			~
			\begin{subfigure}[b]{0.45\textwidth}
				\includegraphics[width=\textwidth]{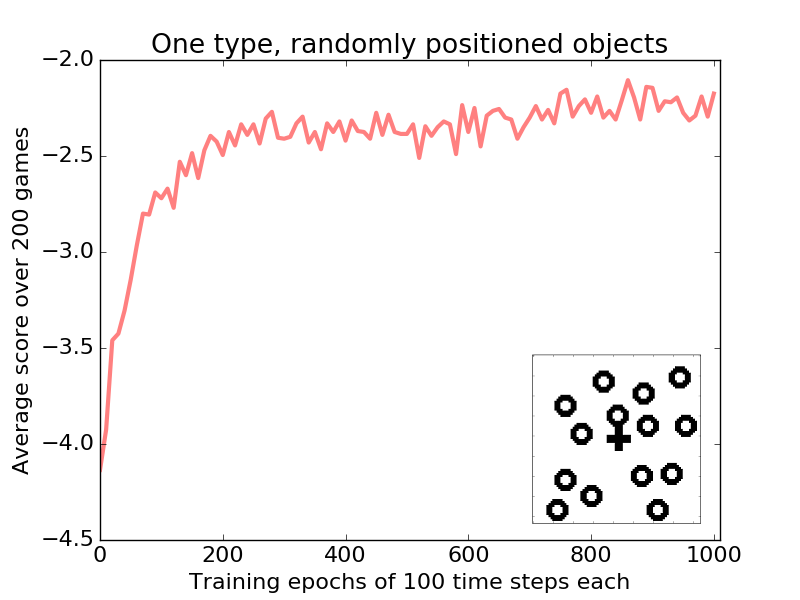}
			\end{subfigure}
			~
			\begin{subfigure}[b]{0.45\textwidth}
				\includegraphics[width=\textwidth]{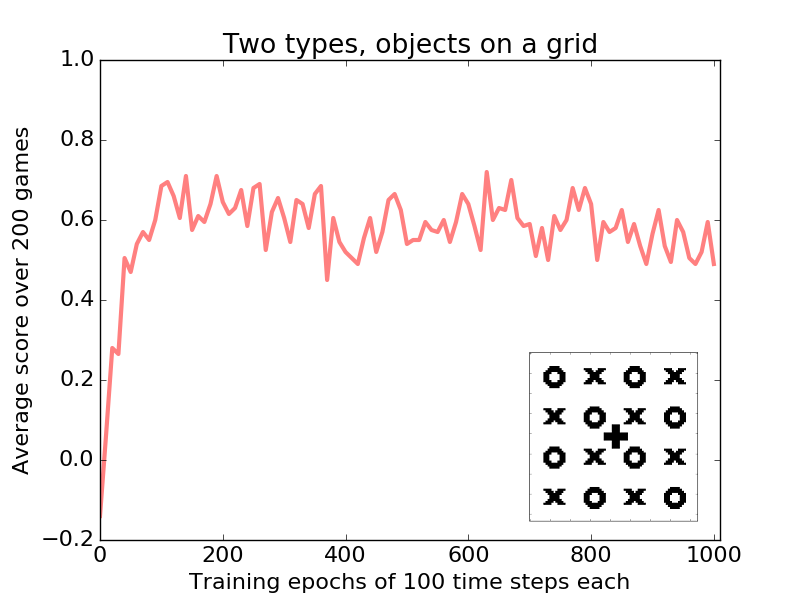}
			\end{subfigure}
			\begin{subfigure}[b]{0.45\textwidth}
				\includegraphics[width=\textwidth]{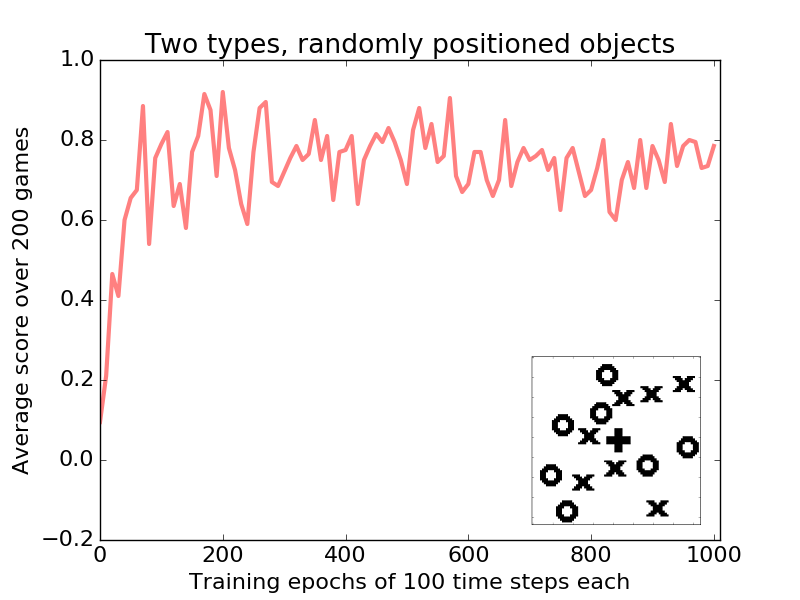}
			\end{subfigure}
			\caption{Average scores for the different game environments}
			\label{fig:LR}
		\end{figure}

		\subsection{Comparison to DQN}
		
		Finally we compare our approach to DQN\footnote{For this comparison we used an open source implementation of DQN: https://github.com/Kaixhin/Atari}. The environments that are suited the best are those with two types of objects as the initial score will be independent of the speed of the agent at the beginning given that they cancel each other out. Figure~\ref{fig:perc} shows the performance of DQN over time. It's important to note that our system's convolutional network was pre-trained on 5000 images, which corresponds to 50 epochs worth of frames. We don't include these in the plots because this pre-training is applicable to all the games and only has to be carried out once.
		
		Thanks to the geometrical simplicity of the grid scenario, the DQN agent quickly learns to move down diagonally to collect only positive objects and avoid negative ones. As a result, the relative number of objects with positive reward for this game variant reaches 100\% after only a few hundred epochs of training, while our agent can only achieve 70\%. On the other hand, when the objects are positioned at random, the DQN agent is not able to learn an effective policy within 1000 epochs, with the number of positive items collected fluctuating around 50\%. So on this game variant, our agent's performance is markedly better than DQN's.
		
		This is also the case for our final experiment where the agent is trained on the grid variant and tested on the random variant (figure~\ref{fig:LR_2}). While our agent rapidly attains a percentage of approximately 70\%, and then fluctuates around that value, DQN is again unable to do better than chance after 1000 epochs. Although we haven't run the experiment long enough to confirm this, we hypothesise that, while DQN might eventually learn to play the random game effectively when trained on the same game, it will never achieve competence at the random game when trained on the grid setup.
		
		\begin{figure}
			\centering
			\begin{subfigure}[b]{0.45\textwidth}
				\includegraphics[width=\textwidth]{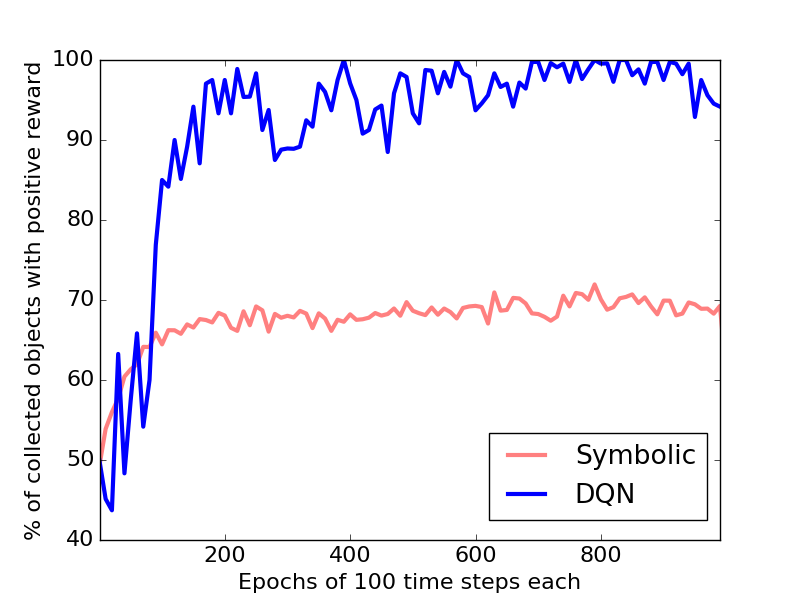}
				\label{fig:percentage_game3}
			\end{subfigure}
			~
			\begin{subfigure}[b]{0.45\textwidth}
				\includegraphics[width=\textwidth]{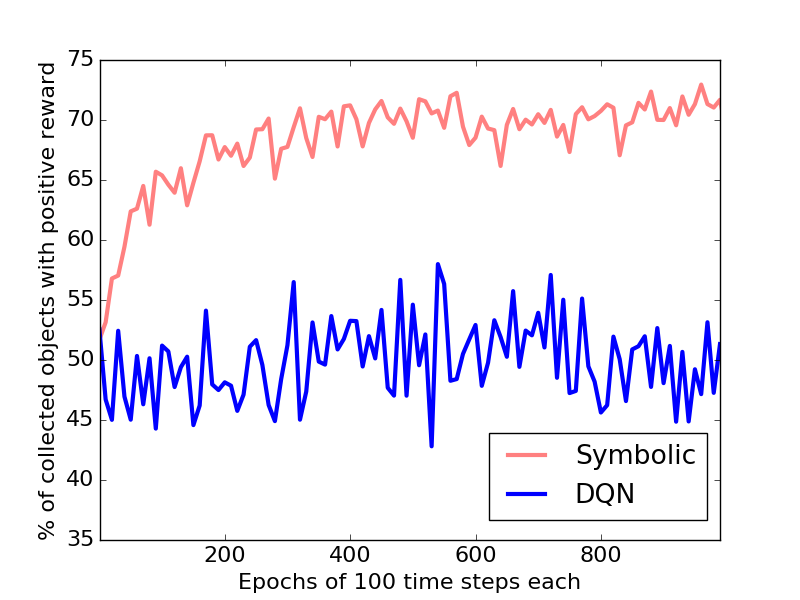}
				\label{fig:percentage_game4}
			\end{subfigure}
			\caption{Comparison between DQN and symbolic approach. Average percentage of objects collected over 200 games that return positive reward in the grid environment (left) and in the random environment (right).}
			\label{fig:perc}
		\end{figure}

		\section{Discussion}
		
		We have proposed a hybrid neural-symbolic, end-to-end reinforcement learning architecture, and claimed that it addresses a number of drawbacks inherent in the current generation of DRL systems. To support this claim, we presented a simple prototype system conforming to the architecture and demonstrated it on several variants of a basic video game. Although the present system cannot learn a globally optimal policy for these games, it learns effectively in all of them. Moreover, even though the system is only a preliminary proof-of-concept with many limitations (to be discussed shortly), it dramatically outperforms DQN on the most difficult game variant, in which the initial placement of objects is random. In this game, DQN's performance didn't exceed chance level after 1000 epochs of training, while our system acquired an effective policy in just 200. We conjecture that DQN struggles with this game because it has to form a statistical picture of all possible object placements, which would require a much larger number of games. In contrast, thanks to the conceptual abstraction made possible by its symbolic front end, our system very quickly ``gets'' the game and forms a set of general rules that covers every possible initial configuration. This demonstration merely hints at the potential for a symbolic front end to promote data efficient learning, potential that we aim to exploit more fully in future work.
		
		Our proof-of-concept system also illustrates one aspect of the architecture's inherent capacity for transfer learning. After training, the unsupervised neural back end of the system is able to form a symbolic representation of any given frame within the micro-world of the game. In effect it has acquired the ontology of that micro-world, and this capability can be applied to any game within that micro-world irrespective of its specific rules. In the present case, no re-training of the back end was required when the system was applied to new variants of the game. However, this form of transfer learning operates at a superficial level. On a more abstract level, the architecture supports far more powerful forms of transfer learning. The key is for the system to understand when a new situation is analogous to one previously encountered or, more potently, to hypothesise that a new situation contains elements of several previously encountered situations combined in a novel way. In the present system, this capability is barely exploited. But in future work we aim to explore the full potential for analogical reasoning made possible by symbolic representation.

		\begin{figure}
			\centering
			\includegraphics[width=0.55\textwidth]{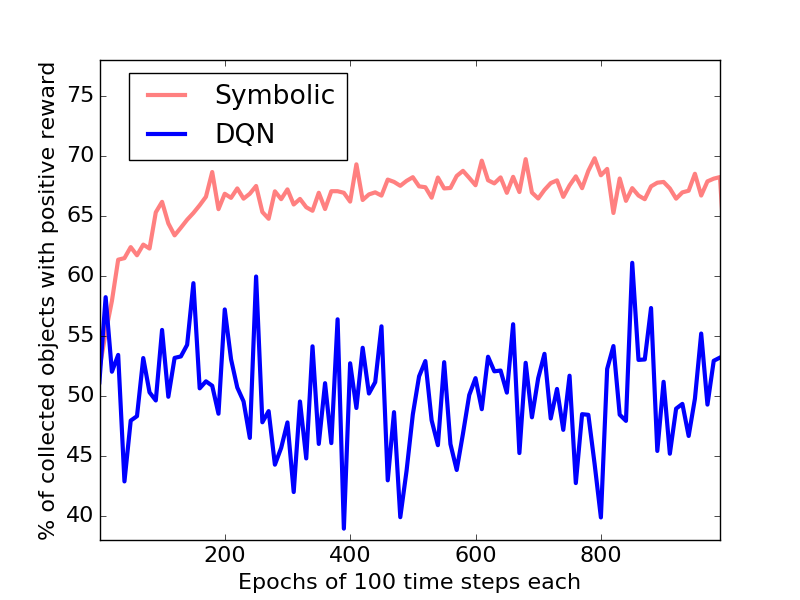}
			\caption{Average percentage of objects collected over 200 games that return positive reward by an agent that is trained on the grid environment and tested on random environments.}
			\label{fig:LR_2}
		\end{figure}

		Finally, because the fundamental mode of operation of the front end of our system is to carry out inference with symbolic representations, there is a humanly-comprehensible chain of justifications for the decisions it makes. For our benchmark example, every action choice can be analysed in terms of the Q functions involved in the decision. Given that these Q functions describe what types of objects are involved in the interaction as well as their relations, we can track back the reasons that led to a certain decision. If the agent chooses to move upwards towards a positive object, for example, we will see that the Q function describing the interaction between this positive object and the agent assigns positive reward to the action of going up. This is a step towards DRL systems with greater transparency.
		
		There are several avenues of further work to explore as we move to more complex games and richer environments with the aim of building a more complete implementation of the deep symbolic reinforcement learning architecture.
		As far as the neural back end is concerned, we intend to make far more extensive use of the state of the art in deep learning. In particular, a more sophisticated deep network capable of unsupervised learning of disentangled representations (having a compositional structure) will be required to handle more realistic images than occur in our benchmark games (eg:~\citep{higgins2016early}).
		
		Similarly, the symbolic back end can draw far more heavily on achievements in classical AI than at present. Three potential elaborations to the architectural blueprint are particularly promising. First, the incorporation of inductive logic programming~\citep{muggleton1991inductive} would enable a more powerful form of generalisation to be applied to the Q function~\citep{dzeroski1998relational}. Recall the claim made in the introduction that general intelligence rests on the ability to determine that a newly encountered situation is similar to one or more situations encountered in the past. Much as a deep neural network is used as a function approximator to achieve this in conventional DRL, inductive logic programming can be used to do this in a symbolic context. Second, to further amplify the system's ability to determine similarity between current and past situations, formal techniques for analogical reasoning can be deployed, such as the structure mapping engine or one of its relatives~\citep{gentner2011analogical}. This technique is especially appropriate when the challenge is not so much to generalise from large numbers of past scenarios, but rather to see that the ongoing situation shares features with a single recorded past episode.
		
		A third way to elaborate the architecture is to build in a planning component that exploits the knowledge of the causal structure of the domain acquired during the learning process. In domains with sparse reward, it's often possible for an agent to discover a sequence of actions leading to a reward state through off-line search rather than on-line exploration. Contemporary logic-based planning methods are capable of efficiently finding large plans in complex domains (eg:\citep{rintanen2012planning}), and it would be rash not to exploit the potential of these techniques. 
		
		Finally, we are by no means ruling out the possibility that the symbolic components of our proposed architecture can themselves be implemented using neural networks. The success of deep learning has inspired a good deal of novel research, much of which involves the innovative use of neural networks with novel architectures. This research could well produce neurally-based implementations of the symbolic reasoning functions we have been advocating here, and there may be advantages to this approach. In the mean time, an architecture that combines deep neural networks with directly implemented symbolic reasoning seems like a promising research direction.

		\section*{Acknowledgments}
		
		We are grateful to Nvidia Corporation for the donation of a high-end GPU. Marta Garnelo is supported by an EPSRC doctoral training award.
		
		\section*{Author Contributions}
		MG \& MS conceptualised the problem and the technical framework. MG developed and tested the algorithms. MG \& MS wrote the paper. MG \& KA carried out comparison with DQN.
		
		\bibliographystyle{plain}
		\bibliography{\jobname}

\begin{thebibliography}{37}
\providecommand{\natexlab}[1]{#1}
\providecommand{\url}[1]{\texttt{#1}}
\expandafter\ifx\csname urlstyle\endcsname\relax
  \providecommand{\doi}[1]{doi: #1}\else
  \providecommand{\doi}{doi: \begingroup \urlstyle{rm}\Url}\fi

\bibitem[LeCun et~al.(2015)LeCun, Bengio, and Hinton]{lecun2015deep}
Yann LeCun, Yoshua Bengio, and Geoffrey Hinton.
\newblock Deep learning.
\newblock \emph{Nature}, 521\penalty0 (7553):\penalty0 436--444, 2015.

\bibitem[Schmidhuber(2015)]{schmidhuber2015deep}
J{\"u}rgen Schmidhuber.
\newblock Deep learning in neural networks: An overview.
\newblock \emph{Neural Networks}, 61:\penalty0 85--117, 2015.

\bibitem[Sutton and Barto(1998)]{sutton1998reinforcement}
Richard~S. Sutton and Andrew~G. Barto.
\newblock \emph{Reinforcement learning: An introduction}.
\newblock MIT press Cambridge, 1998.

\bibitem[Mnih et~al.(2015)Mnih, Kavukcuoglu, Silver, Rusu, Veness, Bellemare,
  Graves, Riedmiller, Fidjeland, Ostrovski, et~al.]{mnih2015human}
Volodymyr Mnih, Koray Kavukcuoglu, David Silver, Andrei~A. Rusu, Joel Veness,
  Marc~G. Bellemare, Alex Graves, Martin Riedmiller, Andreas~K. Fidjeland,
  Georg Ostrovski, et~al.
\newblock Human-level control through deep reinforcement learning.
\newblock \emph{Nature}, 518\penalty0 (7540):\penalty0 529--533, 2015.

\bibitem[Levine et~al.(2016)Levine, Finn, Darrell, and Abbeel]{levine2016end}
Sergey Levine, Chelsea Finn, Trevor Darrell, and Pieter Abbeel.
\newblock End-to-end training of deep visuomotor policies.
\newblock \emph{Journal of Machine Learning Research}, 17\penalty0
  (39):\penalty0 1--40, 2016.

\bibitem[Silver et~al.(2016)Silver, Huang, Maddison, Guez, Sifre, Van
  Den~Driessche, Schrittwieser, Antonoglou, Panneershelvam, Lanctot,
  et~al.]{silver2016mastering}
David Silver, Aja Huang, Chris~J. Maddison, Arthur Guez, Laurent Sifre, George
  Van Den~Driessche, Julian Schrittwieser, Ioannis Antonoglou, Veda
  Panneershelvam, Marc Lanctot, et~al.
\newblock Mastering the game of go with deep neural networks and tree search.
\newblock \emph{Nature}, 529\penalty0 (7587):\penalty0 484--489, 2016.

\bibitem[Legg and Hutter(2007)]{legg2007universal}
Shane Legg and Marcus Hutter.
\newblock Universal intelligence: A definition of machine intelligence.
\newblock \emph{Minds and Machines}, 17\penalty0 (4):\penalty0 391--444, 2007.

\bibitem[Assael et~al.(2015)Assael, Wahlstr{\"{o}}m, Sch{\"{o}}n, and
  Deisenroth]{assael2015data}
John{-}Alexander~M. Assael, Niklas Wahlstr{\"{o}}m, Thomas~B. Sch{\"{o}}n, and
  Marc~Peter Deisenroth.
\newblock Data-efficient learning of feedback policies from image pixels using
  deep dynamical models.
\newblock \emph{arXiv:1510.02173}, 2015.

\bibitem[Gu et~al.(2016)Gu, Lillicrap, Sutskever, and Levine]{gu2016continuous}
Shixiang Gu, Timothy~P. Lillicrap, Ilya Sutskever, and Sergey Levine.
\newblock Continuous deep q-learning with model-based acceleration.
\newblock \emph{arXiv:1603.00748}, 2016.

\bibitem[Parisotto et~al.(2015)Parisotto, Ba, and
  Salakhutdinov]{parisotto2015actor}
Emilio Parisotto, Lei~Jimmy Ba, and Ruslan Salakhutdinov.
\newblock Actor-mimic: Deep multitask and transfer reinforcement learning.
\newblock \emph{arXiv:1511.06342}, 2015.

\bibitem[Arulkumaran et~al.(2016)Arulkumaran, Dilokthanakul, Shanahan, and
  Bharath]{arulkumaran2016classifying}
Kai Arulkumaran, Nat Dilokthanakul, Murray Shanahan, and Anil~Anthony Bharath.
\newblock Classifying options for deep reinforcement learning.
\newblock \emph{arXiv:1604.08153v2}, 2016.

\bibitem[Barreto et~al.(2016)Barreto, Munos, Schaul, and
  Silver]{barreto2016successor}
Andr{\'e} Barreto, R{\'e}mi Munos, Tom Schaul, and David Silver.
\newblock Successor features for transfer in reinforcement learning.
\newblock \emph{arXiv:1606.05312}, 2016.

\bibitem[Rusu et~al.(2016)Rusu, Rabinowitz, Desjardins, Soyer, Kirkpatrick,
  Kavukcuoglu, Pascanu, and Hadsell]{rusu2016progressive}
Andrei~A. Rusu, Neil~C. Rabinowitz, Guillaume Desjardins, Hubert Soyer, James
  Kirkpatrick, Koray Kavukcuoglu, Razvan Pascanu, and Raia Hadsell.
\newblock Progressive neural networks.
\newblock \emph{arXiv:1606.04671}, 2016.

\bibitem[Guo et~al.(2014)Guo, Singh, Lee, Lewis, and Wang]{guo2014deep}
Xiaoxiao Guo, Satinder Singh, Honglak Lee, Richard~L. Lewis, and Xiaoshi Wang.
\newblock Deep learning for real-time atari game play using offline monte-carlo
  tree search planning.
\newblock In \emph{Advances in Neural Information Processing Systems}, pages
  3338--3346, 2014.

\bibitem[Vezhnevets et~al.(2016)Vezhnevets, Mnih, Agapiou, Osindero, Graves,
  Vinyals, and Kavukcuoglu]{mnih2016strategic}
Alexander Vezhnevets, Volodymyr Mnih, John Agapiou, Simon Osindero, Alex
  Graves, Oriol Vinyals, and Koray Kavukcuoglu.
\newblock Strategic attentive writer for learning macro-actions.
\newblock \emph{arXiv:1606.04695}, 2016.

\bibitem[Zahavy et~al.(2016)Zahavy, Ben{-}Zrihem, and
  Mannor]{zahavy2016Graying}
Tom Zahavy, Nir Ben{-}Zrihem, and Shie Mannor.
\newblock Graying the black box: Understanding dqns.
\newblock \emph{arXiv: 1602.02658}, 2016.

\bibitem[McCarthy(1987)]{mccarthy1987generality}
John McCarthy.
\newblock Generality in artificial intelligence.
\newblock \emph{Communications of the ACM}, 30\penalty0 (12):\penalty0
  1030--1035, 1987.

\bibitem[Harnad(1990)]{harnad1990symbol}
Stevan Harnad.
\newblock The symbol grounding problem.
\newblock \emph{Physica D: Nonlinear Phenomena}, 42\penalty0 (1-3):\penalty0
  335--346, 1990.

\bibitem[Shanahan(2005)]{shanahan2005perception}
Murray Shanahan.
\newblock Perception as abduction: Turning sensor data into meaningful
  representation.
\newblock \emph{Cognitive science}, 29\penalty0 (1):\penalty0 103--134, 2005.

\bibitem[Chen et~al.(2016)Chen, Duan, Houthooft, Schulman, Sutskever, and
  Abbeel]{chen2016infogan}
Xi~Chen, Yan Duan, Rein Houthooft, John Schulman, Ilya Sutskever, and Pieter
  Abbeel.
\newblock Infogan: Interpretable representation learning by information
  maximizing generative adversarial nets.
\newblock \emph{arXiv:1606.03657}, 2016.

\bibitem[Goodfellow et~al.(2014)Goodfellow, Pouget-Abadie, Mirza, Xu,
  Warde-Farley, Ozair, Courville, and Bengio]{goodfellow2014generative}
Ian Goodfellow, Jean Pouget-Abadie, Mehdi Mirza, Bing Xu, David Warde-Farley,
  Sherjil Ozair, Aaron Courville, and Yoshua Bengio.
\newblock Generative adversarial nets.
\newblock In \emph{Advances in Neural Information Processing Systems}, pages
  2672--2680, 2014.

\bibitem[Greff et~al.(2016)Greff, Rasmus, Berglund, Hao, Schmidhuber, and
  Valpola]{greff2016tagger}
Klaus Greff, Antti Rasmus, Mathias Berglund, Tele~Hotloo Hao, J{\"{u}}rgen
  Schmidhuber, and Harri Valpola.
\newblock Tagger: Deep unsupervised perceptual grouping.
\newblock \emph{arXiv:1606.06724}, 2016.

\bibitem[Higgins et~al.(2016)Higgins, Matthey, Glorot, Pal, Uria, Blundell,
  Mohamed, and Lerchner]{higgins2016early}
Irina Higgins, Loic Matthey, Xavier Glorot, Arka Pal, Benigno Uria, Charles
  Blundell, Shakir Mohamed, and Alexander Lerchner.
\newblock Early visual concept learning with unsupervised deep learning.
\newblock \emph{arXiv:1606.05579}, 2016.

\bibitem[Kingma and Welling(2013)]{kingma2013auto}
Diederik~P. Kingma and Max Welling.
\newblock Auto-encoding variational bayes.
\newblock \emph{arXiv:1312.6114}, 2013.

\bibitem[Lake et~al.(2016)Lake, Ullman, Tenenbaum, and
  Gershman]{lake2016building}
Brenden~M. Lake, Tomer~D. Ullman, Joshua~B. Tenenbaum, and Samuel~J. Gershman.
\newblock Building machines that learn and think like people.
\newblock \emph{arXiv:1604.00289}, 2016.

\bibitem[Halpern(1990)]{halpern1990analysis}
Joseph~Y. Halpern.
\newblock An analysis of first-order logics of probability.
\newblock \emph{Artificial intelligence}, 46\penalty0 (3):\penalty0 311--350,
  1990.

\bibitem[Bengio et~al.(2013)Bengio, Courville, and
  Vincent]{bengio2013representation}
Yoshua Bengio, Aaron Courville, and Pascal Vincent.
\newblock Representation learning: A review and new perspectives.
\newblock \emph{IEEE transactions on pattern analysis and machine
  intelligence}, 35\penalty0 (8):\penalty0 1798--1828, 2013.

\bibitem[Hayes(1985)]{hayes1985second}
Patrick~J. Hayes.
\newblock The second naive physics manifesto.
\newblock 1985.

\bibitem[Davis(1990)]{davis2014representations}
Ernest Davis.
\newblock \emph{Representations of commonsense knowledge}.
\newblock Morgan Kaufmann, 1990.

\bibitem[Mueller(2014)]{mueller2014commonsense}
Erik~T. Mueller.
\newblock \emph{Commonsense reasoning: An event calculus based approach}.
\newblock Morgan Kaufmann, 2014.

\bibitem[Shanahan(1995)]{shanahan1995default}
Murray Shanahan.
\newblock Default reasoning about spatial occupancy.
\newblock \emph{Artificial Intelligence}, 74\penalty0 (1):\penalty0 147--163,
  1995.

\bibitem[Li et~al.(2016)Li, Liu, Jiang, and Tang]{li2016relief}
Guiying Li, Junlong Liu, Chunhui Jiang, and Ke~Tang.
\newblock Relief impression image detection: Unsupervised extracting objects
  directly from feature arrangements of deep cnn.
\newblock \emph{arXiv:1601.06719}, 2016.

\bibitem[Shanahan(2016)]{shanahan2016frame}
Murray Shanahan.
\newblock The frame problem.
\newblock \emph{The Stanford Encyclopedia of Philosophy}, 2016.

\bibitem[Muggleton(1991)]{muggleton1991inductive}
Stephen Muggleton.
\newblock Inductive logic programming.
\newblock \emph{New generation computing}, 8\penalty0 (4):\penalty0 295--318,
  1991.

\bibitem[D{\v{z}}eroski et~al.(1998)D{\v{z}}eroski, De~Raedt, and
  Blockeel]{dzeroski1998relational}
Sa{\v{s}}o D{\v{z}}eroski, Luc De~Raedt, and Hendrik Blockeel.
\newblock Relational reinforcement learning.
\newblock In \emph{Inductive Logic Programming: 8th International Conference,
  ILP-98 Madison, Wisconsin, USA}, pages 11--22, 1998.

\bibitem[Gentner and Forbus(2011)]{gentner2011analogical}
Dedre Gentner and Kenneth~D. Forbus.
\newblock Computational models of analogy.
\newblock \emph{Wiley interdisciplinary reviews: cognitive science}, 2\penalty0
  (3):\penalty0 266--276, 2011.

\bibitem[Rintanen(2012)]{rintanen2012planning}
Jussi Rintanen.
\newblock Planning as satisfiability: heuristics.
\newblock \emph{Artificial intelligence}, 193:\penalty0 45--86, 2012.

\end{thebibliography}
	\end{document}